\documentclass[letterpaper, 10 pt, conference]{ieeeconf}
\IEEEoverridecommandlockouts
\usepackage{newtxtext}
\usepackage{amsmath,amssymb,amsfonts}
\usepackage{graphicx}
\usepackage{xcolor}
\usepackage{booktabs}
\usepackage{array}
\usepackage{url}
\usepackage[hidelinks]{hyperref}

\graphicspath{{figs/}}

\newcommand{\uniMean}{94.7}
\newcommand{\uniUpMean}{97.0}
\newcommand{\nofallChain}{92.0}
\newcommand{\pushExpZero}{96.9}
\newcommand{\pushDisZero}{81.9}
\newcommand{\pushRLZero}{97.3}
\newcommand{\pushExpForty}{43.9}
\newcommand{\pushDisForty}{41.2}
\newcommand{\pushRLForty}{68.8}

\newcommand{\methodname}{UniExo}
\title{\LARGE\bf \methodname: Unified Multi-Skill Policies for Musculoskeletal\\
Locomotion and Co-Adaptive Exoskeleton Control}

\author{Yifei~Yuan,
        Jakob~Wolf,
        Ghaith~Androwis,
        and~Xianlian~Zhou%
\thanks{Yifei Yuan, Jakob Wolf and Xianlian Zhou are with the Department of Biomedical
Engineering, New Jersey Institute of Technology, Newark, NJ 07102, USA
(e-mail: yy72@njit.edu; jw768@njit.edu; alexzhou@njit.edu).}%
\thanks{Ghaith Androwis is with Kessler Foundation, West Orange, NJ 07052, USA
(e-mail: GAndrowis@kesslerfoundation.org).}%
}

\begin{document}
\maketitle
\thispagestyle{empty}
\pagestyle{empty}

%==============================================================================
\begin{abstract}
Daily locomotion encompasses diverse activities and frequent transitions between them, yet most exoskeleton controllers are designed for a single activity or a narrow set of related movements. Changes in activity therefore typically require explicit mode switching and separately tuned or retrained controllers. Simulation-based learning reduces the need for hardware-based tuning but generally retains this limitation. Here we present \methodname{}, a framework that first constructs a multi-skill musculoskeletal human policy and then jointly trains an exoskeleton control policy with it. Four single-skill imitation experts for walking, turning, running and backward walking are distilled into a single network structured by a skill latent and subsequently fine-tuned through reinforcement learning on transition sequences. The resultant unified human policy achieves a mean tracking success rate of \uniMean\% on unseen clips of the four skills and exhibits greater robustness to perturbations than its constituent experts. 
A single hip exoskeleton controller (\methodname{}) is initialized from hip moment prediction of the human policy and co-adapted with it through multi-agent reinforcement learning across the four skills. This co-adaptation shifts the timing of the assistance torque and raises the fraction of positive work delivered to the hip. When deployed on a custom hip exoskeleton, the controller generalizes across four treadmill speeds in six participants and assists one participant through a continuous route of all four skills and their transitions, without skill labels or explicit mode switching. 
\methodname{} thus provides a step towards replacing activity-specific controllers with unified, user-specific controllers that support diverse locomotor activities and the transitions between them.
\end{abstract}

%==============================================================================
\section{Introduction}

Lower-limb exoskeletons provide a promising approach to rehabilitation and mobility augmentation. Exoskeleton controllers have been optimized for level walking through human-in-the-loop experiments \cite{zhang2017human,ding2018human,slade2022personalizing} and parametrized for walking and running \cite{lim2023parametric}. Task-agnostic control from an estimate of the biological joint moment \cite{molinaro2024task} needs no activity-specific design but relies on a large laboratory dataset of labeled joint moments and has to be retrained for each new activity. Learning in simulation makes controller development experiment-free \cite{luo2023robust,luo2024experiment,leem2026exo}, and controllers have been learned this way for walking, running, stair climbing and sit-to-stand \cite{luo2024experiment,ratnakumar2026predicting}. A co-adaptive controller trained in simulation has also been validated on participants with metabolic and biomechanical measurements \cite{yuan2026staged}. Across all of this work the controller is still built around a single activity or a fixed set of activities, and none so far handles a range of activities together with the transitions between them.

Closing this gap calls for a human model that can produce a range of activities and the transitions between them, from which an exoskeleton controller can then be learned in simulation and without hardware tuning. 
The simulated human is a muscle-driven musculoskeletal model with far more actuators than degrees of freedom. OpenSim set the model-based workflow for musculoskeletal analysis \cite{delp2007opensim}. MyoSuite brought the same models into MuJoCo as contact-rich RL environments \cite{caggiano2022myosuite}, and MyoAssist composes them with assistive-device models \cite{robbins2026myoassist}. MS-Human-700 scales the model to 700 muscle-tendon units and controls it through a learned low-dimensional action space \cite{zuo2024self}. Making these models move requires a control policy, usually learned by imitation from motion capture. Training the human together with the device, as in \cite{luo2024experiment,yuan2026smat}, couples two policies that adapt to each other, and the human policy in these works is trained on a single gait. KINESIS \cite{simos2025kinesis} and MuscleMimic \cite{li2026towards} both train one musculoskeletal policy on large motion datasets and reproduce a wide range of motions with human-like muscle activation, but the policy in each case is evaluated on imitation alone and has not been coupled with an assistive device. A multi-skill human policy that closes the loop with an exoskeleton controller is still missing.
Extending co-adaptation beyond a single gait is not straightforward, since the shared assistance controller must now generalize across skills with different cadence and loading, and the controller must infer which skill is underway without an explicit label, as standing, turning and running demand markedly different assistance from the same hip actuator. 

\begin{figure*}[!t]
\centering
\includegraphics[width=\textwidth]{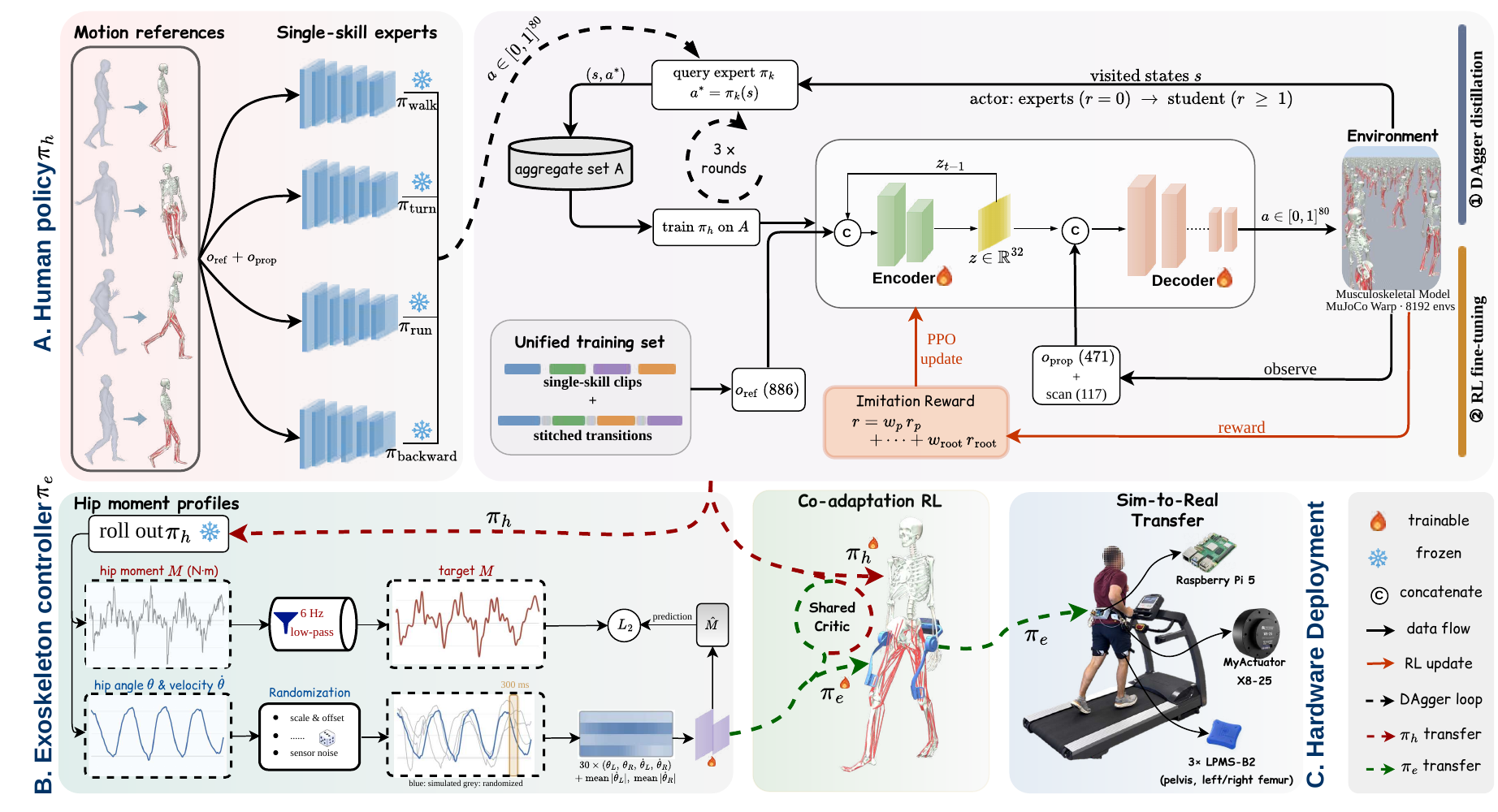}
\vspace{-18pt}
\caption{Overview of \methodname{}. \textbf{(A) Human policy $\pi_h$.} Four single-skill experts are trained by imitation and frozen. \textcircled{1} DAgger distillation: the experts label the states visited by the student, the labeled states are added to the aggregate set $A$ and the student is retrained on $A$ for three rounds, which gives the distilled policy. The encoder maps the reference motion $o_{\text{ref}}$ and the previous latent $z_{t-1}$ to the skill latent $z$ and the decoder maps $z$ and proprioception $o_{\text{prop}}$ to muscle activations $a$. \textcircled{2} RL fine-tuning: the distilled policy is fine-tuned with PPO on single-skill clips and stitched transitions, which gives the unified policy $\pi_h$. \textbf{(B) Exoskeleton controller $\pi_e$.} Rollouts of the frozen $\pi_h$ provide the hip moment and hip kinematics of each skill. A predictor learns the low-pass filtered moment from a $300$~ms window of randomized hip kinematics and initializes $\pi_e$. $\pi_h$ and $\pi_e$ are then co-adapted with PPO and a shared critic. \textbf{(C) Hardware deployment.} $\pi_e$ runs on a Raspberry~Pi~5 with three LPMS-B2 IMUs and drives two MyActuator X8-25 motors.}
\label{fig:pipeline}
\end{figure*}

We present \methodname{}, which first builds a multi-skill human policy on MyoLeg, the 80-muscle lower-limb model of MyoSuite. The policy is trained by distilling four single-skill experts into one network organized around a low-dimensional skill latent and fine-tuning it with reinforcement learning on transition sequences. An exoskeleton controller is then initialized from the hip moment of this policy and co-adapted with it to learn an assistive torque profile before deployment on a custom hip exoskeleton (Fig.~\ref{fig:pipeline}). This work makes three main contributions. First, we show that single-skill imitation experts can be distilled into one latent-conditioned policy and fine-tuned on stitched transition sequences, giving a single musculoskeletal controller that performs multiple locomotion skills and moves between them without a mode signal (Sec.~\ref{sec:human}). Second, we extend human--exoskeleton co-adaptation to a multi-skill setting, using one shared-weight bilateral controller and domain randomization of the input kinematics as its only input (Sec.~\ref{sec:exo}). Third, we validate this pipeline on hardware without per-user retuning, across a range of treadmill speeds and along a continuous multi-skill route (Sec.~\ref{sec:hardware}--\ref{sec:results}).
\emph{The code will be made publicly available.}

%==============================================================================

\section{Musculoskeletal Simulation and Imitation Learning}

\subsection{Models and simulation framework}
\label{sec:sim} 

We use MyoLeg, a biologically accurate lower-limb musculoskeletal model from MyoSuite \cite{caggiano2022myosuite} built on the MuJoCo physics simulator \cite{todorov2012mujoco}. The model actuates the legs with 80 musculotendon units and represents the upper body as a rigid body. Its knee follows the OpenSim convention \cite{delp2007opensim}, in which several auxiliary coordinates are bound to the knee flexion angle by polynomial constraints. The exoskeleton is modeled from its CAD geometry and coupled to MyoLeg, with the belt rigidly tied to the pelvis and each flexion axis coaxial with the human hip. The thigh cuffs are attached through compliant sliding constraints, because the exoskeleton abduction axis does not coincide with that of the hip and a rigid attachment would lock ab/adduction. The exoskeleton provides assistance to the hip joint through an additional actuator group.

We pair the MuJoCo Warp physics simulation engine with \emph{mjlab} \cite{zakka2026mjlab}, a manager-based environment layer built on the Isaac~Lab API \cite{mittal2023orbit} for reinforcement learning. The Isaac~Lab ray-cast height sensor is used for terrain perception, though every result here is on flat ground. The human policy action is an 80-dimensional muscle excitation vector mapped to activations in $[0,1]$, and the exoskeleton controller action is a two-dimensional torque command, one per hip, with maximum torque $\tau_{\max}$. During simulation, physics steps at 300~Hz and the policy acts at $100$~Hz. We trained both policies with PPO from \emph{rsl\_rl} \cite{rudin2022learning} on a single NVIDIA RTX~PRO~6000 Blackwell GPU, with 8192 parallel environments for the human policy and 2048 for co-adaptation, sustaining $3.7\times10^{4}$ environment steps per second including the learning update.

\subsection{Reference motion and retargeting}
\label{sec:retarget}
The human control policy is trained through imitation learning from a large set of motions that is summarized in Table~\ref{tab:data}. It includes 920 clips and 81.1~min of locomotion, all of which were resampled to $100$~Hz. Walk, turn and backward-walk clips are from AMASS \cite{mahmood2019amass}. Run clips are from LAFAN1 \cite{harvey2020robust}, taken from the MyoSkeleton retargeting released with LocoMuJoCo \cite{al2023locomujoco}. Each skill is split into training and test clips in an approximate 9:1 ratio, with both sets covering a similar range of movement directions and clip lengths.
The walk, turn and backward clips come as SMPL \cite{loper2023smpl} joint positions and the run clips as MyoSkeleton joint angles, which already include all MyoLeg joints. Both sources are converted to two kinds of reference. The first is the world position of seven tracked bodies, the pelvis and the left and right tibia, talus and toes, used by the imitation reward and the termination criterion. The second is the 28-joint configuration of MyoLeg, used to set the pose at reset and in the joint-angle reward. For the SMPL clips, inverse kinematics (IK) matches the pelvis and the four foot bodies but not the two knee positions: in our IK fits, matching the SMPL knee position, which carries a bend absent from the source motion, produced a visible crouch. We therefore take the knee flexion angle from SMPL directly instead. The tracked-body positions are then recomputed from the IK joint angles with the SMPL knee angle by forward kinematics, so that the body-position targets are kinematically consistent with the joint-angle targets. For the MyoSkeleton clips the joint angles are used directly and the positions come from forward kinematics.

\begin{table}[t]
\vspace*{5pt}
\caption{Training and Test Motion Data of the Four Skills}
\label{tab:data}
\centering\footnotesize
\begin{tabular*}{\columnwidth}{@{\extracolsep{\fill}}lrrrrr@{}}
\toprule
 & Walk & Turn & Backward & Run & All\\
\midrule
Train clips   & 428  & 144  & 108 & 147  & 827\\
Train minutes & 36.2 & 16.2 & 9.6 & 10.8 & 72.8\\
Test clips    & 48   & 19   & 10  & 16   & 93\\
Test minutes  & 4.1  & 2.1  & 0.9 & 1.2  & 8.3\\
\bottomrule
\end{tabular*}\\[3pt]
\parbox{\columnwidth}{\footnotesize\raggedright Transition sequences: 600 (129.8 min), built from the training clips.
Unified policy training set: 827 clips + 600 sequences = 1427 (202.6 min).}
\end{table}

\subsection{Imitation task}
\label{sec:task}
The human muscle control policy observes 1357 values. Proprioception accounts for 471 of them: foot contacts (4), root height and tilt (5), body poses and velocities in the root frame (222), tendon lengths and velocities (160) and actuator forces (80). The remaining 886 carry the reference motion: the tracking error (84), the reference pose in the root frame (21), five future reference frames sampled every $0.20$~s (700), the motion phase (1) and the previous action (80).
% Future TODO: state what each of the five future reference frames contains (i.e., the composition of the 140 values per frame) for reproducibility.

Let $p_i,v_i$ be the position and velocity of tracked body $i\in\{1,\dots,7\}$, with $i{=}1$ the pelvis, let $q_j,\dot q_j$ be the 28 joint angles and velocities, and let $a\in[0,1]^{80}$ be the muscle activations, with reference quantities carrying a hat. Define the pelvis-referenced errors $\tilde p_i=(\hat p_i-\hat p_1)-(p_i-p_1)$ and $\tilde v_i$ likewise, and the joint errors $\tilde q_j=\hat q_j-q_j$ and $\dot{\tilde q}_j$. Four imitation terms $r_k$, $k=1,\dots,4$, share one form,
\begin{equation}
r_k=\exp\!\left(-\alpha_k\,\overline{e_k^2}\right),
\label{eq:imit}
\end{equation}
where $e_k$ is the term's error and $\overline{\;\cdot\;}$ averages it over the seven bodies or the 28 joints: body position uses $\tilde p_i$ with $(w_k,\alpha_k)=(0.5,200)$, body velocity $\tilde v_i$ with $(0.05,5)$, joint angle $\tilde q_j$ with $(0.4,20)$ and joint velocity $\dot{\tilde q}_j$ with $(0.05,1)$. This exponential form is standard in physics-based imitation \cite{peng2018deepmimic} and the decomposition into position, velocity and joint terms follows \cite{li2026towards}. Here the body terms are pelvis-referenced and constrain posture alone, while the joint terms constrain the full joint configuration and enforce the knee correction of Sec.~\ref{sec:retarget} during training. Besides these imitation reward terms, three additional reward terms are used:
\begin{align}
r_{\text{up}} &= \exp\!\big(-3\,(\theta_x^2+\theta_y^2)\big), & w_{\text{up}}&=0.1,\\
r_{\text{eng}} &= \exp\!\big(-0.05\,(\lVert a\rVert_1+\lVert a\rVert_2)\big), & w_{\text{eng}}&=0.05,\\
r_{\text{root}} &= \exp\!\big(-5\,\lVert \hat v_1^{xy}-v_1^{xy}\rVert^2\big), & w_{\text{root}}&=0.2,
\end{align}
where $\theta_x$ and $\theta_y$ are the forward and sideways tilt angles of the pelvis from vertical and $\lVert a\rVert_1$ and $\lVert a\rVert_2$ are the L1 and L2 norms of the muscle activations. The three terms keep the pelvis upright, penalize muscle effort and match the pelvis velocity to the reference in the world frame, and $r_{\text{root}}$ is added only when fine-tuning the unified policy. 
The total reward is thus defined as:
\begin{equation}
r=\sum_{k=1}^{4} w_k r_k + w_{\text{up}}r_{\text{up}} + w_{\text{eng}}r_{\text{eng}} + w_{\text{root}}r_{\text{root}}.
\label{eq:total_r}
\end{equation}
An episode ends when the mean pelvis-referenced deviation of the seven tracked bodies from the reference exceeds a fixed threshold of $0.3$~m.

%==============================================================================
\section{Unified Multi-Skill Human Policy}
\label{sec:human}

\subsection{Single-skill experts}
Musculoskeletal models have far more muscle actuators than degrees of freedom, which makes accurate imitation over a wide range of motions difficult to learn, so we first train one expert for each skill on its clips in Table~\ref{tab:data}.
The walk, turn, backward-walk and run experts are trained from scratch with PPO \cite{schulman2017proximal,rudin2022learning}. All four share an identical actor--critic: a $2048$--$1536$--$1024$--$1024$--$512$--$512$ MLP with LayerNorm and SiLU, sized as in \cite{simos2025kinesis}. PPO uses $8192\times50$ steps per iteration, 10 epochs, 4 minibatches, learning rate $5\times10^{-5}$ adapted to a target KL of $0.01$, $\gamma{=}0.99$, $\lambda{=}0.95$, clip $0.2$ and entropy coefficient $0$. Each expert trains for $2.9$--$4.9\times10^{9}$ environment steps, or 21.7--36.9~h.

\subsection{Unified latent architecture}
\label{sec:latent}
To learn the four skills in a single network and to achieve smooth transitions between them, an encoder--latent--decoder network is used for the unified human policy $\pi_h$ (Fig.~\ref{fig:pipeline}),
\begin{equation}
z_t \sim \mathrm{enc}\!\left(o^{\text{ref}}_t,\, z_{t-1}\right), \qquad a_t = \mathrm{dec}\!\left(o^{\text{prop}}_t,\, o^{\text{scan}}_t,\, z_t\right),
\label{eq:enc}
\end{equation}
where $o^{\text{ref}}\!\in\!\mathbb{R}^{886}$ is the reference observation, $o^{\text{prop}}\!\in\!\mathbb{R}^{471}$ the proprioception, $o^{\text{scan}}\!\in\!\mathbb{R}^{117}$ a terrain height scan ($13\times9$ grid) with $0.1$~m resolution, $z\!\in\!\mathbb{R}^{32}$ the skill latent and $a\!\in\![0,1]^{80}$ the muscle activations. The encoder is a $1024$--$512$ MLP whose 64 outputs are the mean and the standard deviation of the Gaussian over $z_t$. Its input also carries $z_{t-1}$, so each latent continues from the one before. The decoder has the same layer sizes as an expert. The encoder receives the reference motion, the decoder receives the body state, and the target motion reaches the muscles through $z$. 
Terrain enters at the decoder as a height (distance) scan relative to the pelvis, and the flat ground provides an input of 117 values all equal to the pelvis height.

\begin{figure}[t]
\centering
\includegraphics[width=\columnwidth]{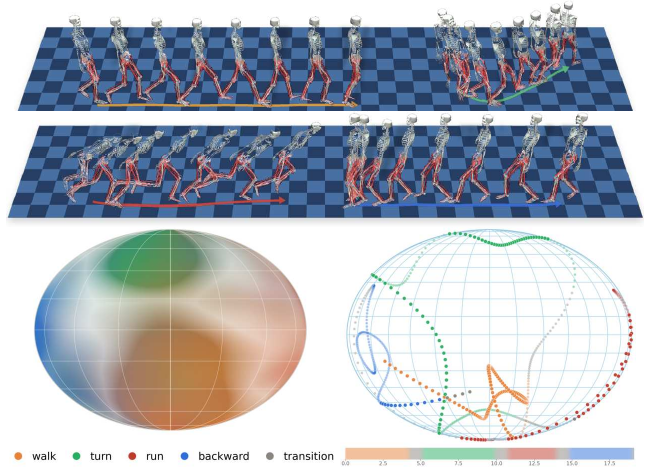}
\vspace{-18pt}
\caption{Skill latent $z$ of the unified policy. Top: a sequence of walk, turn, run and backward motions. Left: the motion sequence in the discriminant projection, normalized onto a sphere. Right: one stitched walk--turn--run--backward sequence in the same view, colored by time.}
\label{fig:zspace}
\vspace{-7pt}
\end{figure}

\subsection{Distilled policy for multiple activities}
\label{sec:dagger}

The four experts are distilled into the unified policy with DAgger \cite{ross2011reduction}, which reduces the compounding error of behavior cloning by labeling states sampled from the state distribution of the student with the actions of the corresponding expert. In each round the expert acts with probability $\beta$ and the student with probability $1-\beta$, and the labeled states are appended to the aggregate set $\mathcal{D}_r$ on which the student is retrained,
\begin{multline}
\pi_h \leftarrow \arg\min_{\theta}\ \mathbb{E}_{\mathcal{D}_r} \Big[\big\lVert \pi_\theta(s_t)-\pi^{k}(s_t)\big\rVert^{2}\\ + \lambda_{\text{KL}}\,\mathrm{KL}\Big(\mathrm{enc}\big(o^{\text{ref}}_t, z_{t-1}\big)\,\Big\|\,\mathcal{N}\big(0.95\,z_{t-1},\,(1-0.95^{2})I\big)\Big)\Big],
\label{eq:bc}
\end{multline}
where $\pi^{k}$ is the expert for the skill being tracked, the expectation runs over the 30-step sequences of $\mathcal{D}_r$, and the KL term with $\lambda_{\text{KL}}=0.001$ keeps successive latents correlated. Round~0 uses $\beta{=}1$ because no student has been trained yet and later rounds use $\beta{=}0$, with each round appending $4\times8192\times30$ state--action pairs over the four skills to $\mathcal{D}_r$. Fine-tuning starts from the round-3 student because it reaches the highest success rate under action noise.

\subsection{Unified policy fine-tuning with transition sequences}
\label{sec:unified}
The distilled policy is fine-tuned with PPO on the unified-policy training set of Table~\ref{tab:data} to recover the tracking accuracy of the experts and to learn the transitions between skills that the single-skill expert data does not contain. The single-skill clips preserve the four skills during fine-tuning and the policy learns to switch between skills from transition sequences that we build by stitching training clips through a standing pose. Each sequence concatenates 2--4 clips sampled from the training set. Each clip is translated in the horizontal plane and rotated about the vertical axis so that its initial pelvis position and heading coincide with the final pelvis position and heading of the preceding clip, and consecutive clips are connected by a $1.5$~s standing pose. The pelvis position is continuous across each junction to within 1~cm. Fine-tuning runs for 8000 iterations in 26.7~h with the early-termination threshold of Sec.~\ref{sec:task} fixed at $0.3$~m and the expert hyperparameters unchanged. 
Fig.~\ref{fig:zspace} shows the skill latent of the resulting policy, projected by linear discriminant analysis onto the three directions that best separate the four skills. The four skills occupy separate regions of the latent space and a stitched sequence traverses them continuously.

%==============================================================================
\section{Exoskeleton Controller}
\label{sec:exo}

The exoskeleton controller $\pi_e$ for the bilateral hip exoskeleton of Sec.~\ref{sec:sim} is built in two stages (Fig.~\ref{fig:pipeline}B): a network is trained on rollouts of the unified policy to predict the hip moment from hip kinematics, and the predictor is then fine-tuned by co-adaptation with the human policy.

\subsection{Stage 1: kinematic hip moment predictor}

\label{sec:stage2}
We roll out the unified policy on the four skills and record the hip kinematics and the sagittal hip moment of both legs at every step, and a $256$--$128$ MLP with SiLU is trained on these rollouts to predict the hip moment from hip kinematics. Its input is a stacked $300$~ms history at $100$~Hz: the hip angle and angular velocity of both legs over that window plus each leg's mean angular velocity $|\omega|$ over the window as a measure of hip motion intensity, giving 122 inputs. $|\omega|$ is appended here to signal the intensity of hip movement and clearly differentiate activities such as standing and other activities. For the policy output, instead of using this network to predict moments for both legs, we chose to use the same network separately for each leg, predicting joint moment per leg. For each leg the input is assembled with that leg first and the other leg second, and the network is run once per leg with shared weights. This ensures symmetry in the predicted moments when the kinematics is symmetric. 

To generalize across users and to bridge the gap between simulated and measured kinematics, we apply domain randomization to the input kinematics while the target is unchanged. The randomization warps the time axis of each rollout by about $30\%$, scales and offsets the hip angle, shifts the other leg in time with a standard deviation of $15$~ms, adds a velocity bump with a standard deviation of $0.35$~rad/s in stance, and adds Gaussian sensor noise to the angle and velocity. The bump imitates a velocity bump in mid-stance that we observed on the thigh IMU in the experiments and that may be caused by soft-tissue motion or sensor placement. Targets are low-pass filtered with a 6~Hz Butterworth and divided by the peak moment of their skill, and the predictor is trained to minimize the mean squared error between its output $\hat M$ and this normalized moment over both legs and all steps. With many input variants mapped to the same filtered moment at each step, the network converges to the moment shape shared across strides.

\begin{table*}[t]
\vspace*{5pt}
\caption{Policy Performance Across the Three Training Stages}
\label{tab:success}
\centering
\renewcommand{\arraystretch}{1.15}
\footnotesize
\begin{tabular*}{\textwidth}{@{\extracolsep{\fill}}l|cccc|cccc|cccc@{}}
\hline
 & \multicolumn{4}{c|}{Single-skill experts} & \multicolumn{4}{c|}{Distilled (DAgger)} & \multicolumn{4}{c}{Unified} \\
\cline{2-5}\cline{6-9}\cline{10-13}
Skill & Track. & Upright & Cov. & MPJPE & Track. & Upright & Cov. & MPJPE & Track. & Upright & Cov. & MPJPE \\
\hline
Walk     & 95.8 & 93.8 & 97.8 & 28.9 & 87.7 & 85.1 & 93.5 & 34.3 & 93.2 & 94.7 & 94.2 & 29.2 \\
Turn     & 96.0 & 94.8 & 99.1 & 24.7 & 64.4 & 64.6 & 87.3 & 39.5 & 99.7 & 99.7 & 99.5 & 30.7 \\
Backward & 99.9 & 99.8 & 100.0 & 26.3 & 89.1 & 76.4 & 93.4 & 40.0 & 92.1 & 94.7 & 83.4 & 31.0 \\
Run      & 100.0 & 93.8 & 100.0 & 29.0 & 96.1 & 97.7 & 96.3 & 33.4 & 93.8 & 98.9 & 95.9 & 33.2 \\[2pt]
\hline\\[-9pt]
\textbf{Mean} & \textbf{97.9} & \textbf{95.5} & \textbf{99.2} & \textbf{27.2} & \textbf{84.3} & \textbf{81.0} & \textbf{92.6} & \textbf{36.8} & \textbf{\uniMean} & \textbf{\uniUpMean} & \textbf{93.2} & \textbf{31.0} \\
\hline
\end{tabular*}\\[2pt]
\parbox{\textwidth}{\scriptsize\raggedright Track., Upright and Cov. in \%, MPJPE in mm. Track.: episodes whose global tracked-body error, the world-frame position error averaged over the seven tracked bodies, stays below $0.5$~m. Upright: deviation does not terminate and an episode counts only if the pelvis stays above $0.80$~m throughout. Cov.: fraction of the clip completed before the global tracked-body error exceeds $0.3$~m. MPJPE: per-joint position error after pelvis alignment.}
\end{table*}

\subsection{Stage 2: human--exoskeleton co-adaptation}

\label{sec:stage3}
The predictor reproduces the hip moment of the unassisted human, but this moment applied as torque does not necessarily follow the direction of hip motion or provide optimal assistance. This stage therefore co-adapts the exoskeleton with the human to learn the timing of the torque relative to hip motion, since this timing differs between skills \cite{lim2023parametric}. Both policies are actors that share one critic and one PPO update as in \cite{yuan2026smat}. $\pi_e$ observes the same $300$~ms window for each leg and is initialized from the Stage-1 predictor, and then maps its output to the commanded torque
\begin{equation}
\tau_{\text{exo}}=\tau_{\max}\,\mathrm{clip}\!\left(\hat M,-1,+1\right)\ \xrightarrow{\ \text{6\,Hz low-pass}\ }\ \tau_{\text{cmd}},
\end{equation}
where $\tau_{\max}$ is $15$~N$\cdot$m in training. 
%The $0.3$~m deviation threshold is unchanged. 
In addition to the human reward (Eq.~\ref{eq:total_r}), the exoskeleton reward has four terms:
\begin{align}
r_{\text{exo}} =\;& 1.0\underbrace{\textstyle\sum_j (0.3\,\bar\tau_j\bar\omega_j - 0.15\,\bar\tau_j^2)}_{\text{assistance power}}
 + 3.0\underbrace{e^{-0.025\,|\tau_{\text{muscle}}|}}_{\text{net-moment offload}}\nonumber\\
 &+ 1.0\underbrace{e^{-4\,\bar a_{\text{hip}}}}_{\text{muscle saving}}
 - 4.0\underbrace{\|\bar\tau_t-\bar\tau_{t-1}\|^2}_{\text{smoothness}} ,
\label{eq:exoreward}
\end{align}
where $\bar\tau_j=\tau_{\text{exo},j}/\tau_{\max}$ is the normalized command of leg $j$, $\bar\omega_j$ its hip angular velocity scaled to $[-1,1]$, $\tau_{\text{muscle}}=\tau_{\text{net}}-\tau_{\text{exo}}$ the muscle's own share of the sagittal hip moment averaged over both legs, and $\bar a_{\text{hip}}$ the mean activation of the 22 sagittal hip muscles. Assistance power is the mechanical power the exoskeleton delivers to the joint, regularized by $\bar\tau^2$ against saturation. Net-moment offload is the main objective. Motion imitation or tracking closely regulates $\tau_{\text{net}}$, so the muscle's share of it falls only when the device takes over part of the load. Muscle saving guards against co-contraction, which offload alone would not detect. Smoothness penalizes rapid changes in the delivered torque.
And the total reward during this co-adaptation is the sum of the human reward (Eq. \ref{eq:total_r}) and the exoskeleton reward (Eq. \ref{eq:exoreward}).

\section{Sim-to-Real Policy Deployment}
\label{sec:hardware}

\subsection{Device design and electronics}
A bilateral hip exoskeleton was used to test the trained \methodname{} controller. The device is machined in aluminum and weighs $4.23$~kg (Fig.~\ref{fig:pipeline}C). Each side has one active degree of freedom in the sagittal plane, driven by a MyActuator~X8-25 brushless motor with a peak torque of $25$~N$\cdot$m, and one passive degree of freedom in ab/adduction just below the motor. A Raspberry~Pi~5 runs the controller and logs at $100$~Hz, reaching the two actuators over separate CAN buses and three LPMS-B2 inertial measurement units (IMUs) over Bluetooth Low Energy. The hip flexion angle and angular velocity of each leg come from the thigh unit relative to the pelvis unit and pass through a One Euro filter.

\subsection{Experimental design}
Two experiments were run on the device with $\tau_{\max}=15$~N$\cdot$m. The first experiment tests speed generalization. Six participants (age $23.3\pm3.3$~yr, height $176\pm4$~cm, mass $77.5\pm9.3$~kg) walked on a treadmill at $0.76$, $1.25$, $1.74$ and $2.25$~m/s under two conditions, with and without assistance. IMU kinematics and the commanded torque were logged at $100$~Hz. Assistance is scored by the positive-work fraction
\begin{equation}
\eta^{+} \;=\; \frac{\sum_t \max\!\left(W_t,\,0\right)}{\sum_t \lvert W_t\rvert}\ \in[0,1],
\label{eq:poswork}
\end{equation}
with per-step work $W_t=\tau_{\text{exo},t}\,\omega_t\,\Delta t$ summed over both legs.
%, which equals $1$ when the device only assists the joint and $0$ when it only resists it. 
The second experiment tests the controller across skills. One of the participants followed a clockwise route on the laboratory floor, walking, turning, running and walking backward, repeated five times. The study had IRB approval (Protocol No.\ 2305033091R003) with written informed consent from all participants.

%==============================================================================
\section{Results}
\label{sec:results}

\begin{figure}[t]
\centering
\includegraphics[width=\columnwidth]{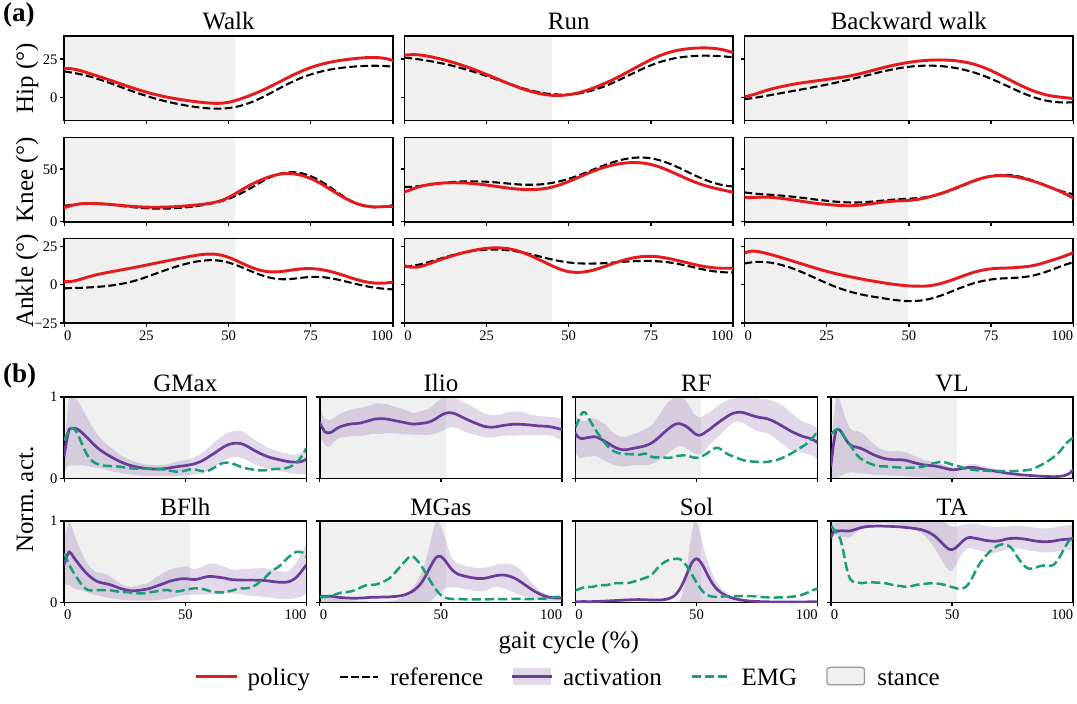}
\vspace{-18pt}
\caption{Gait cycle of the unified policy, averaged over cycles segmented at heel strike. (a) Hip, knee and ankle angles for walking, running and backward walking: policy (red) and reference (black dashed), stance shaded. (b) Muscle activation during walking, normalized per muscle, with surface EMG (green dashed). Gluteus maximus (GMax), iliopsoas (Ilio), rectus femoris (RF), vastus lateralis (VL), biceps femoris long head (BFlh), gastrocnemius medialis (MGas), soleus (Sol) and tibialis anterior (TA).}
\label{fig:gaitmuscle}
\end{figure}

\subsection{Evaluation of the unified human policy in simulation}

Table~\ref{tab:success} reports the performance of the human policies from the three stages of Sec.~\ref{sec:human} on the test clips of Table~\ref{tab:data}. The single-skill experts reach a mean tracking success rate $3.2\%$ higher than the unified policy, but each covers one skill and cannot switch to another. Fine-tuning raises the tracking success rate of the distilled policy by $10.4\%$ and also makes the unified policy more stable than the experts, where stability is measured by the upright rate, the fraction of episodes run without deviation termination in which the pelvis stays above $0.80$~m. The unified policy has the highest upright rate of the three stages on three of the four skills.

When the unified policy exceeds the $0.5$~m tracking threshold, it mostly does not fall. Backward walking is the clearest case: its pelvis-aligned joint error stays at $31$~mm, within $5$~mm of the expert, while the world-frame error grows to $101$~mm. Therefore, the gait is intact and the model drifts in position late in the clip. The same policy walks backward upright on $94.7\%$ of those clips. On the stitched multi-skill sequences it stays on its feet through \nofallChain\% of episodes.

\textbf{Kinematics and muscle activation.} Fig.~\ref{fig:gaitmuscle} averages joint angles and muscle activation over gait cycles segmented at heel strike, with the backward cycle starting at forefoot contact because the foot lands toe first. Stance covers $53$, $45$ and $51\%$ of the cycle for walking, running and backward walking. Hip, knee and ankle follow the reference motion on all three skills, with the hip and ankle $5$--$7^\circ$ more flexed than the reference and the knee peak in swing $6^\circ$ lower during running ($56$ against $62^\circ$). Muscle activation over the walking cycle is overlaid with surface EMG, from \cite{schreiber2019multimodal} for gluteus maximus and \cite{camargo2021comprehensive} for the others. Iliopsoas has no surface EMG because it is a deep muscle. The EMG is scaled to the peak of the simulated mean to compare activation timing. Gluteus maximus, vastus lateralis and biceps femoris long head peak within $5\%$ of the cycle of the EMG, gastrocnemius medialis and soleus peak $8$--$11\%$ later, and rectus femoris peaks in swing where the EMG peaks at loading.

\begin{figure}[t]
\centering
\includegraphics[width=\columnwidth]{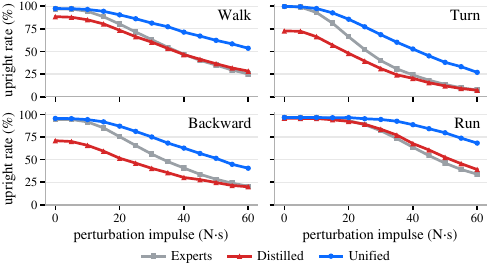}
\vspace{-18pt}
\caption{Upright rate under pelvis pushes for the three training stages: the
single-skill experts, the distilled policy and the unified policy.}
\label{fig:push}
\end{figure}

\textbf{Robustness to perturbation.} Distillation can only reproduce the experts on the states they visit, so a perturbation outside the training distribution is what separates the unified policy from the experts it was distilled from: a $0.25$~s horizontal random push applied at a height of $0.90$~m on the pelvis every $2$~s at a random gait phase. Fig.~\ref{fig:push} shows the upright rate of the three stages under impulses from $0$ to $60$~N$\cdot$s. Without a push, the unified policy is as stable as the experts, \pushRLZero\% upright against \pushExpZero\%, while the distilled policy it started from stays at \pushDisZero\%. The difference shows under larger impulses, where at $40$~N$\cdot$s the unified policy stays upright on \pushRLForty\% of episodes, the experts on \pushExpForty\% and the distilled policy on \pushDisForty\%. Fine-tuning therefore makes the policy more robust to disturbance than the experts.

\subsection{Exoskeleton controller}
\label{sec:exoresults}

\begin{figure}[t]
\centering
\includegraphics[width=\columnwidth]{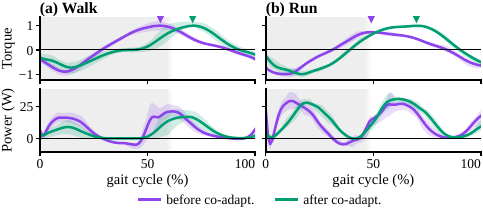}
\vspace{-18pt}
\caption{Simulated exoskeleton torque and power over the gait cycle for (a) walking and (b) running, right leg, before and after co-adaptation. Torque is normalized to its peak and the triangles mark the peaks. Stance is shaded.}
\label{fig:exocycle}
\end{figure}

\begin{table}[t]
\caption{Exoskeleton Controller Before and After Co-Adaptation}
\label{tab:coadapt}
\centering\footnotesize
\setlength{\tabcolsep}{3pt}
\begin{tabular*}{\columnwidth}{@{\extracolsep{\fill}}lrrrr@{}}
\toprule
& $P^{+}/P^{-}$, S1 & \textbf{$P^{+}/P^{-}$, S2} & $\eta^{+}$, S1 & \textbf{$\eta^{+}$, S2}\\
\midrule
Walk     & 7.0/1.4 & \textbf{6.0/0.4} & 82.9 & \textbf{94.4}\\
Turn     & 3.4/0.9 & \textbf{3.3/0.2} & 79.7 & \textbf{95.3}\\
Backward & 3.3/2.5 & \textbf{3.7/0.8} & 57.1 & \textbf{82.3}\\
Run      & 9.3/2.1 & \textbf{11.3/0.7} & 81.8 & \textbf{94.0}\\
\bottomrule
\end{tabular*}\\[2pt]
\parbox{\columnwidth}{\scriptsize\raggedright S1/S2: Stage~1/Stage~2. $P^{+}/P^{-}$ (W): mean positive/negative power of the right leg. $\eta^{+}$ (\%): positive-work fraction. Evaluated on the unified-policy rollouts.}
\end{table}

\begin{figure}[t]
\centering
\includegraphics[width=\columnwidth]{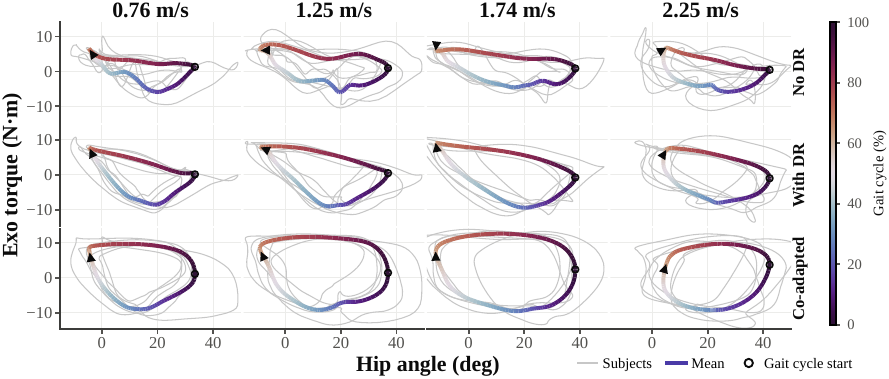}
\vspace{-18pt}
\caption{Exoskeleton work loops at four treadmill speeds from the same recorded hip kinematics. Rows: predictor without domain randomization (DR), with DR, and the co-adapted controller on the device. Thin lines are participants and the thick line the mean, colored by gait-cycle phase.}
\label{fig:exp1speed}
\end{figure}

Table~\ref{tab:coadapt} evaluates the exoskeleton controller before and after co-adaptation. Both stages receive the hip kinematics of the unified-policy rollouts as input. The difference between the two stages is the torque profile. Negative power falls by $65$--$81\%$ on the four skills while positive power changes by $-15$ to $+22\%$. $\eta^{+}$ rises accordingly, by $11.5\%$ on walking, $15.6\%$ on turning, $25.2\%$ on backward walking and $12.2\%$ on running. Fig.~\ref{fig:exocycle} shows the torque profile before and after co-adaptation for walking and running. The flexion peak arrives $10\%$ of the cycle later on walking and $9\%$ later on running and moves into early swing. Power stays positive over a larger part of the cycle on both skills.

\begin{figure}[t]
\centering
\includegraphics[width=\columnwidth]{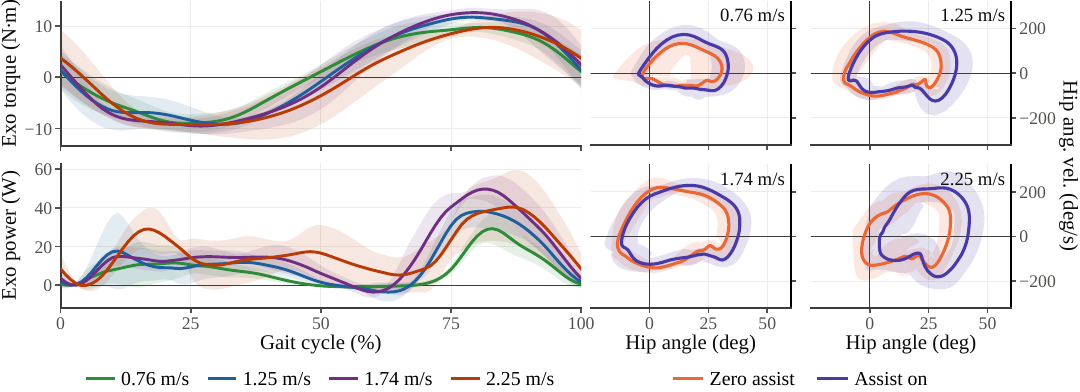}
\vspace{-18pt}
\caption{Speed generalization on the treadmill. Left: exoskeleton torque and power over the gait cycle at four speeds. Right: hip angle against hip angular velocity with zero assist and with assist on. Lines are the six-participant mean and bands one standard deviation.}
\label{fig:exp1profile}
\end{figure}

\begin{figure*}[t]
\centering
\includegraphics[width=\textwidth]{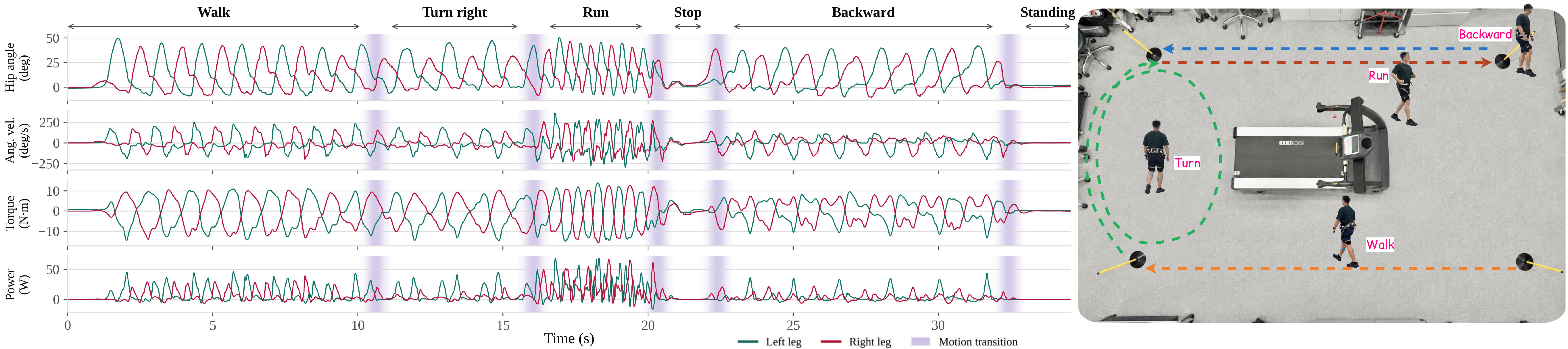}
\vspace{-18pt}
\caption{One participant on the laboratory route with \methodname{}. Left: hip angle, hip angular velocity, exoskeleton torque and power of both legs over one repetition. Bands mark transitions between motions. Right: the route and the order of motions.}
\label{fig:exp2}
\end{figure*}

\subsection{Speed generalization}
Six participants walked on a treadmill at $0.76$, $1.25$ and $1.74$~m/s and jogged at $2.25$~m/s, with and without assistance from the \methodname{} controller. 
%The same controller serves all four speeds. 
Fig.~\ref{fig:exp1speed} plots exoskeleton torque against hip angle over the gait cycle for three policies: the Stage-1 trained policies with and without domain randomization and the final co-adapted \methodname{} controller. Without randomization the loops are noisy and inconsistent between participants. With randomization all four speeds demonstrate smoother and consistent patterns. The final \methodname{} controller enlarges the loop further, indicating more net work per stride, and moves the peak torque later in the cycle. 

Fig.~\ref{fig:exp1profile} shows the torque and power profiles and the hip kinematics. The torque keeps one shape across speeds with a flexion peak in late stance that grows from $10.6$~N$\cdot$m at $0.76$~m/s to $13.4$~N$\cdot$m at $1.74$~m/s and falls to $11.7$~N$\cdot$m at $2.25$~m/s.
Positive work is $96.2\pm2.1\%$ of the total and net work per stride is $9.7$ to $15.8$~J per leg across speeds. Compared to walking without assistance, assistance expands the phase portrait at every speed and the hip range of motion grows by $3$ to $5^\circ$. 

\subsection{Assistance across skills and transitions}
Fig.~\ref{fig:exp2} shows the results for one participant walking on the multi-skill route in clockwise direction with the \methodname{} controller. The participant walked, turned right, ran, stopped, walked backward and stood still. The torque remains smooth through the transitions without jumps at the change of skill, and during standing it remains nearly constant with near-zero power, so the controller does not need to recognize the activity or switch modes. The torque adapts its period and amplitude to each skill. The flexion peak is highest in running at $14.1$~N$\cdot$m and lowest in walking backward at $10.2$~N$\cdot$m, and the extension peak stays close to the $15$~N$\cdot$m limit in the three forward skills, averaged over the five repetitions. When the participant starts running the torque period shortens from $1.4$ to $0.7$~s within the first stride. Positive work is $94$, $96$, $98$ and $90\%$ of the total in the four skills. 

%==============================================================================
\section{Discussion}

Unlike prior musculoskeletal policies trained on large motion datasets, which are evaluated on imitation alone \cite{simos2025kinesis,li2026towards}, the unified policy here is coupled to a downstream task: it supplies both the training partner and the hip-moment target for the exoskeleton controller. The robustness gained through fine-tuning (Sec.~\ref{sec:results}) is what makes this coupling viable, since perturbations and user variability push the system outside the states the experts were distilled on. The same distillation-and-fine-tuning recipe should extend to further skills without architectural changes.

\methodname{} learns one exoskeleton controller for four skills and the transitions between them. The controller applies one network with shared weights to each leg independently, so the policy treats the two legs symmetrically. Three design choices support the transfer of the exoskeleton controller from simulation to hardware. Domain randomization of the input kinematics in Stage~1 maps many variants of the hip kinematics to the same filtered moment so that the predictor learns the moment shape shared across strides, and without it the predictor produces noisy work loops that differ between participants on the recorded kinematics (Fig.~\ref{fig:exp1speed}). The same randomization covers the differences between users and between the simulated model and the IMU measurements, and one controller assisted six participants without per-user tuning. The smoothness term of Eq.~\ref{eq:exoreward} penalizes changes in the commanded torque between steps during co-adaptation. Low-pass filtering of the command at $6$~Hz and of the IMU signals with a One Euro filter removes the remaining high-frequency content on the device. Together these gave a torque that follows the motion without oscillation on hardware.

The hardware experiments validate the controller output but do not yet establish its physiological effect on the user. The torque profile, the positive-work fraction and the hip kinematics were measured across speeds on six participants on the treadmill and across skills on one participant overground, and metabolic cost was not measured.
Several limitations also remain in simulation. The human policy is limited to four locomotion skills on flat ground and does not include activities such as stair climbing, slope walking or sit-to-stand. The simulated exoskeleton is also idealized, since its belt is rigidly tied to the pelvis and its flexion axes are coaxial with the hip, whereas on the device the belt and cuffs move with the soft tissue and the flexion axes can be misaligned with the hip. Its muscle activation is not fully consistent with experimentally observed EMG patterns; for example, rectus femoris peaks in swing in the simulation, whereas EMG shows a peak during loading. The muscle-saving term of Eq.~\ref{eq:exoreward} covers the sagittal hip muscles alone, so co-contraction outside this set is not penalized and a reduction in hip activation may shift load to other muscles.

\section{Conclusion}
We have shown that a single unified musculoskeletal policy, distilled from single-skill experts and fine-tuned on transition sequences, can support multiple locomotor skills and the transitions between them, and that co-adapting an exoskeleton controller (\methodname{}) with this policy transfers to hardware without per-user retuning. This points to a general path for building assistive controllers around unified, user-specific human policies rather than around single activities. Future work will extend the framework to a broader range of daily activities and evaluate its physiological benefits, including metabolic cost, in larger cohorts.

{\footnotesize
\bibliographystyle{IEEEtran}
\bibliography{references}
}

\end{document}